\documentclass[10pt,twocolumn,letterpaper]{article}

\usepackage{cvpr}
\usepackage{times}
\usepackage{epsfig}
\usepackage{graphicx}
\usepackage{amsmath}
\usepackage{amssymb}

\usepackage{multirow}
\usepackage{mathtools}

\usepackage{caption} 
\usepackage{makecell}
\usepackage{pgfplots}
\pgfplotsset{compat = 1.10}
\usepgfplotslibrary{external} 
\usepackage{float}
\usepackage{subfig}
\usepackage{graphicx}

\definecolor{cyan}{RGB}{35,164,205}
\definecolor{pinegreen}{RGB}{65,151,24}
\definecolor{gold}{RGB}{253,161,59}
\definecolor{yellow}{RGB}{252,231,36}
\definecolor{green}{RGB}{114,207,85}
\definecolor{teal}{RGB}{54,90,140}
\definecolor{purple}{RGB}{70,9,92}

\pgfplotsset{
    colormap={colorbrewer-ylgnbu}{[1pt]
        color(0pt)=(purple);
        color(17pt)=(teal);
        color(35pt)=(green);
        color(50pt)=(yellow);
    },
}

\usepackage[breaklinks=true,bookmarks=false]{hyperref}

\cvprfinalcopy 

\def\cvprPaperID{3990} 

\begin{document}

\title{Shift: A Zero FLOP, Zero Parameter Alternative to Spatial Convolutions}

\author{Bichen Wu, Alvin Wan\thanks{Authors contributed equally.}, Xiangyu Yue$^*$, Peter Jin, Sicheng Zhao, \\Noah Golmant, Amir Gholaminejad, Joseph Gonzalez, Kurt Keutzer \\
UC Berkeley\\
{\tt\small \{bichen,alvinwan,xyyue,phj,schzhao,noah.golmant,amirgh,jegonzal,keutzer\}@berkeley.edu}
}

\maketitle

\begin{abstract}
Neural networks rely on convolutions to aggregate spatial information. However, spatial convolutions are expensive in terms of model size and computation, both of which grow quadratically with respect to kernel size. In this paper, we present a parameter-free, FLOP-free ``shift'' operation as an alternative to spatial convolutions. We fuse shifts and point-wise convolutions to construct end-to-end trainable shift-based modules, with a hyperparameter characterizing the tradeoff between accuracy and efficiency. To demonstrate the operation's efficacy, we replace ResNet's 3x3 convolutions with shift-based modules for improved CIFAR10 and CIFAR100 accuracy using 60\% fewer parameters; we additionally demonstrate the operation's resilience to parameter reduction on ImageNet, outperforming ResNet family members. We finally show the shift operation's applicability across domains, achieving strong performance with fewer parameters on classification, face verification and style transfer.

\end{abstract}
\section{Introduction and Related Work}

Convolutional neural networks (CNNs) are ubiquitous in computer vision tasks, including image classification, object detection, face recognition, and style transfer. These tasks enable many emerging mobile applications and Internet-of-Things (IOT) devices; however, such devices have significant memory constraints and restrictions on the size of over-the-air updates (e.g. 100-150MB). This in turn imposes constraints on the size of the CNNs used in these applications. For this reason, we focus on reducing CNN model size while retaining accuracy, on applicable tasks. 

\begin{figure}[ht]
\centering
\includegraphics[width=0.47\textwidth]{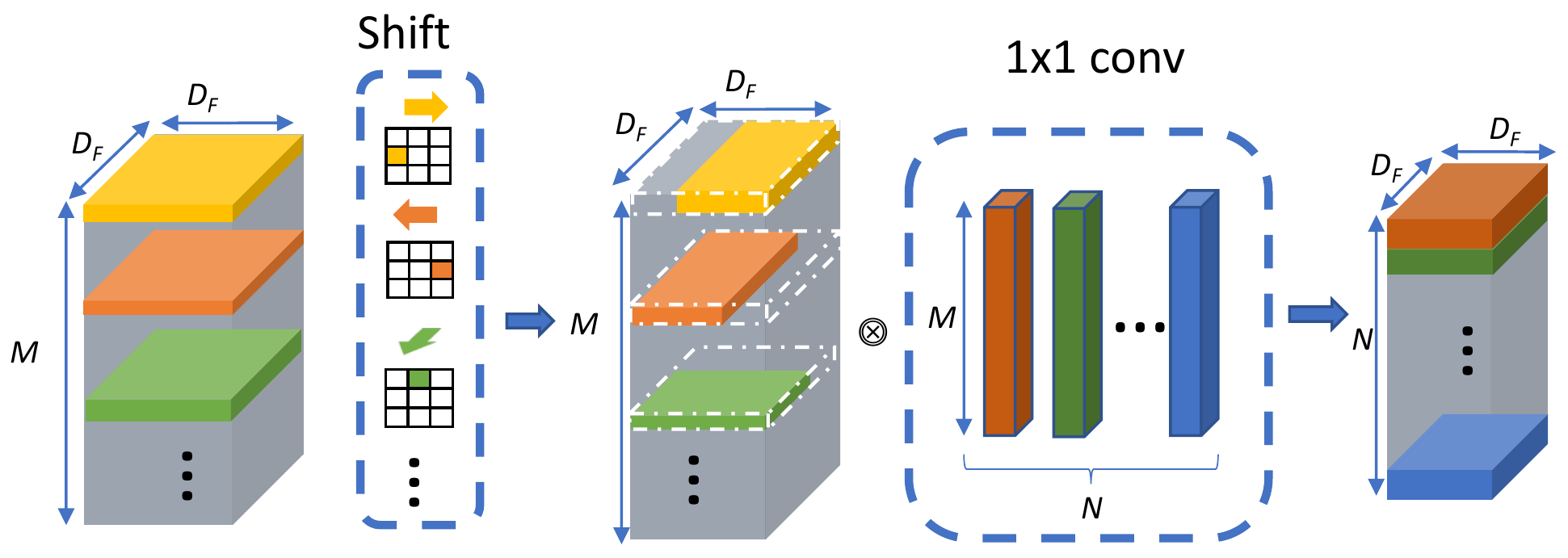}
   \caption{Illustration of a shift operation followed by a 1x1 convolution. The shift operation adjusts data spatially and the 1x1 convolution mixes information across channels.}
\label{fig:shift-conv}
\end{figure}

CNNs rely on spatial convolutions with kernel sizes of 3x3 or larger to aggregate spatial information within an image. However, spatial convolutions are very expensive in both computation and model size, each of which grows quadratically with respect to kernel size. In the VGG-16 model \cite{VGG}, 3x3 convolutions account for 15 million parameters, and the \textit{fc1} layer, effectively a 7x7 convolution, accounts for 102 million parameters.

Several strategies have been adopted to reduce the size of spatial convolutions. ResNet\cite{ResNet} employs a ``bottleneck module,'' placing two 1x1 convolutions before and after a 3x3 convolution, reducing its number of input and output channels. Despite this, 3x3 convolutional layers still account for  50\% of all parameters in ResNet models with bottleneck modules. SqueezeNet \cite{SqueezeNet} adopts a ``fire module,'' where the outputs of a 3x3 convolution and a 1x1 convolution are concatenated along the channel dimension. 
Recent networks such as ResNext~\cite{RexNext}, MobileNet~\cite{MobileNet}, and Xception~\cite{Xception} adopt group convolutions and depth-wise separable convolutions as alternatives to standard spatial convolutions. In theory, depth-wise convolutions require less computation. However, it is difficult to implement depth-wise convolutions efficiently in practice, as their arithmetic intensity (ratio of FLOPs to memory accesses) is too low to efficiently utilize hardware.
Such a drawback is also mentioned in \cite{ShuffleNet, Xception}. ShuffleNet~\cite{ShuffleNet} integrates depth-wise convolutions, point-wise group convolutions, and channel-wise shuffling to further reduce parameters and complexity. In another work, \cite{LocalBinaryCNN} inherits the idea of a separable convolution to freeze spatial convolutions and learn only point-wise convolutions. This does reduce the number of learnable parameters but falls short of saving FLOPs or model size. 

\begin{figure*}[ht!]
\centering
\includegraphics[width=1.\textwidth]{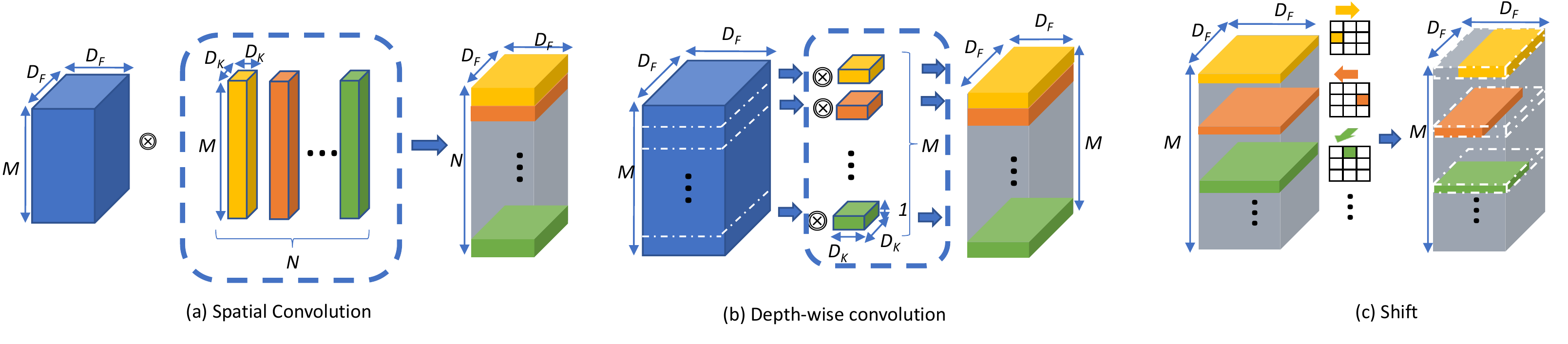}
   \caption{Illustration of (a) spatial convolutions, (b) depth-wise convolutions and (c) shift. In (c), the 3x3 grids denote a shift matrix with a kernel size of 3. The lighted cell denotes a 1 at that position and white cells denote 0s.}
\label{fig:conv-depth-shift}
\vspace{-0.2in}
\end{figure*}

Our approach is to sidestep spatial convolutions entirely. In this paper, we present the \textit{shift operation} (Figure~\ref{fig:shift-conv}) as an alternative to spatial convolutions. 
The shift operation moves each channel of its input tensor in a different spatial direction. A \textit{shift-based module} interleaves shift operations with point-wise convolutions, which further mixes spatial information across channels. 
Unlike spatial convolutions, the shift operation itself requires zero FLOPs and zero parameters. As opposed to depth-wise convolutions, shift operations can be easily and efficiently implemented.

Our approach is orthogonal to model compression~\cite{DeepCompression}, tensor factorization~\cite{TensorFact} and low-bit networks~\cite{XnorNet}. As a result, any of these techniques could be composed with our proposed method to further reduce model size.

We introduce a new hyperparameter for shift-based modules, ``expansion'' $\mathcal{E}$, corresponding to the tradeoff between FLOPs/parameters and accuracy. This allows practitioners to select a model according to specific device or application requirements.  Using shift-based modules, we then propose a new family of architectures called \textit{ShiftNet}. To demonstrate the efficacy of this new operation, we evaluate ShiftNet on several tasks: image classification, face verification, and style transfer. Using significantly fewer parameters, ShiftNet attains competitive performance.

\section{The Shift Module and Network Design}

We first review the standard spatial and depth-wise convolutions illustrated in Figure~\ref{fig:conv-depth-shift}. Consider the spatial convolution in Figure~\ref{fig:conv-depth-shift}(a), which takes a tensor $F \in \mathbb{R}^{D_F \times D_F \times M}$ as input. Let $D_F$ denote the height and width and $M$ denote the channel size. The kernel of a spatial convolution is a tensor $K\in \mathbb{R}^{D_K \times D_K \times M \times N}$, where $D_K$ denotes the kernel's spatial height and width, and $N$ is the number of filters. For simplicity, we assume the stride is 1 and that the input/output have identical spatial dimensions. Then, the spatial convolution outputs a tensor $G \in \mathbb{R}^{D_F\times D_F \times N}$, which can be computed as
\begin{equation}
    G_{k, l, n} = \sum_{i, j, m}K_{i, j, m, n} F_{k+\hat{i},l+\hat{j}, m},
\end{equation}
where $\hat{i} = i - \lfloor D_F/2\rfloor, \hat{j} = j - \lfloor D_F/2 \rfloor$ are the re-centered spatial indices; $k, l$ and $i, j$ index along spatial dimensions and $n, m$ index into channels.
The number of parameters required by a spatial convolution is $M\times N\times D_K^2$ and the computational cost is $M\times N\times D_K^2 \times D_F^2 $. As the kernel size $D_K$ increases, we see the number of parameters and computational cost grow quadratically. 

A popular variant of the spatial convolution is a depth-wise  convolution \cite{MobileNet,Xception}, which is usually followed by a point-wise convolution (1x1 convolution). Altogether, the module is called the depth-wise separable convolution.
A depth-wise convolution, as shown in Figure~\ref{fig:conv-depth-shift}(b), aggregates spatial information from a $D_K \times D_K$ patch within each channel, and can be described as 
\begin{equation}
    \hat{G}_{k, l, m} = \sum_{i, j}\hat{K}_{i, j, m} F_{k+\hat{i},l+\hat{j}, m},
\end{equation}
where $\hat{K} \in \mathbb{R}^{D_F\times D_F\times M}$ is the depth-wise convolution kernel. This convolution comprises $M\times D_K^2$ parameters and $M \times D_K^2 \times D_F^2$ FLOPs. As in standard spatial convolutions, the number of parameters and computational cost grow quadratically with respect to the kernel size $D_K$. Finally, point-wise convolutions mix information across channels, giving us the following output tensor 
\begin{equation}
    G_{k, l, n} = \sum_{m}P_{m, n} \hat{G}_{k,l, m},
    \label{eqn:point-wise}
\end{equation}
where $P\in\mathbb{R}^{M\times N}$ is the point-wise convolution kernel. 

In theory, depth-wise convolution requires less computation and fewer parameters. In practice, however, this means memory access dominates computation, thereby limiting the use of parallel hardware. 
For standard convolutions, the ratio between computation \textit{vs.} memory access is 
\begin{equation}
    \frac{M\times N \times D_F^2 \times D_K^2}{D_F^2\times (M+N) + D_K^2\times M \times N},
    \label{eqn:conv-AI}
\end{equation}
while for depth-wise convolutions, the ratio is 
\begin{equation}
    \frac{M\times D_F^2 \times D_K^2}{D_F^2\times 2M + D_K^2\times M}.
    \label{eqn:depth-AI}
\end{equation}
A lower ratio here means that more time is spent on memory accesses, which are several orders of magnitude slower and more energy-consuming than FLOPs. 
This drawback implies an I/O-bound device will be unable to achieve maximum computational efficiency.

\subsection{The Shift Operation}

The shift operation, as illustrated in Figure~\ref{fig:conv-depth-shift}(c), can be viewed as a special case of depth-wise convolutions. Specifically, it can be described logically as:
\begin{equation}
    \tilde{G}_{k, l, m} = \sum_{i, j}\tilde{K}_{i, j, m} F_{k+\hat{i},l+\hat{j}, m}.
    \label{eqn:shift}
\end{equation}
The kernel of the shift operation is a tensor $\tilde{K} \in \mathbb{R}^{D_F \times D_F\times M}$ such that 
\begin{equation}
    \tilde{K}_{i, j, m} = 
    \begin{cases}
        1, & \text{if } i=i_m \text{ and } j=j_m,  \\
        0, & \text{otherwise}.
    \end{cases}
\end{equation}
Here $i_m, j_m$ are channel-dependent indices that assign one of the values in $\tilde{K}_{:, :, m} \in \mathbb{R}^{D_K\times D_K}$ to be 1 and the rest to be 0. We call $\tilde{K}_{:, :, m}$ a shift matrix. 

For a shift operation with kernel size $D_K$, there exist $D_K^2$ possible shift matrices, each of them corresponding to a shift direction. If the channel size $M$ is no smaller than $D_K^2$, we can construct a shift matrix that allows each output position $(k, l)$ access to all values within a $D_K \times D_K$ window in the input. We can then apply another point-wise convolution per Eq.~(\ref{eqn:point-wise}) to exchange information across channels. 

Unlike spatial and depth-wise convolutions, the shift operation itself does not require parameters or floating point operations (FLOPs). Instead, it is a series of memory operations that adjusts channels of the input tensor in certain directions. A more sophisticated implementation can fuse the shift operation with the following 1x1 convolution, where the 1x1 convolution directly fetches data from the shifted address in cache. With such an implementation, we can aggregate spatial information using shift operations, for free.

\subsection{Constructing Shift Kernels}

For a given kernel size $D_K$ and channel size $M$, there exists $D_K^2$ possible shift directions, making $(D_K^2)^M$ possible shift kernels. An exhaustive search over this state space for the optimal shift kernel is prohibitively expensive. 

To reduce the state space, we use a simple heuristic: divide the $M$ channels evenly into $D_K^2$ groups, where each group of $\lfloor M/D_K^2 \rfloor$ channels adopts one shift. We will refer to all channels with the same shift as a \textit{shift group}. The remaining channels are assigned to the ``center'' group and are not shifted.

However, finding the optimal permutation, \textit{i.e.}, how to map each channel-$m$ to a shift group, requires searching a combinatorially large search space. To address this issue, we introduce a modification to the shift operation that makes input and output invariant to channel order: We denote a shift operation with channel permutation $\pi$ as $\mathcal{K}_{\pi}(\cdot)$, so we can express Eq.~(\ref{eqn:shift}) as $\tilde{G} = \mathcal{K}_{\pi}(F)$. We permute the input and output of the shift operation as  
\begin{equation}
\tilde{G} = \mathcal{P}_{\pi_2}(\mathcal{K}_{\pi}(\mathcal{P}_{\pi_1}(F))) = (\mathcal{P}_{\pi_2}\circ \mathcal{K}_{\pi} \circ \mathcal{P}_{\pi_1}) (F),
\label{eqn:perm-shift-perm}
\end{equation}
where $\mathcal{P}_{\pi_i}$ are permutation operators and $\circ$ denotes operator composition. However, permutation operators are discrete and therefore difficult to optimize. As a result, we process the input $F$ to Eq.~(\ref{eqn:perm-shift-perm}) by a point-wise convolution $\mathcal{P}_1 (F)$. We repeat the process for the output $\tilde{G}$ using $\mathcal{P}_2 (\tilde{G})$. The final expression can be written as 
\begin{equation}
\begin{aligned}
    G & = (\mathcal{P}_2 \circ \mathcal{P}_{\pi_2} \circ \mathcal{K}_{\pi} \circ \mathcal{P}_{\pi_1} \circ \mathcal{P}_1 ) (F) \\
    & = ((\mathcal{P}_2 \circ \mathcal{P}_{\pi_2}) \circ \mathcal{K}_{\pi} \circ (\mathcal{P}_{\pi_1} \circ \mathcal{P}_1) ) (F) \\ 
    & = (\hat{\mathcal{P}}_2 \circ \mathcal{K}_{\pi} \circ \hat{\mathcal{P}}_1) (F), 
\end{aligned}
\label{eqn:SCS-operation}
\end{equation}
where the final step holds, as there exists a bijection from the set of all $\mathcal{P}_i$ to the set of all $\mathcal{\hat{P}}_i$, since the permutation operator $\mathcal{P}_{\pi_i}$ is bijective by construction. As a result, it suffices to learn $\hat{\mathcal{P}}_1$ and $\hat{\mathcal{P}}_2$ directly. Therefore, this augmented shift operation Eq.~(\ref{eqn:SCS-operation}) can be trained with stochastic gradient descent end-to-end, without regard for channel order. So long as the shift operation is sandwiched between two point-wise convolutions, different permutations of shifts are equivalent. Thus, we can choose an arbitrary permutation for the shift kernel, after fixing the number of channels for each shift direction.

\subsection{Shift-based Modules}

First, we define a \textit{module} to be a collection of layers that perform a single function, e.g.~ResNet's bottleneck \textit{module} or SqueezeNet's fire \textit{module}. Then, we define a \textit{group} to be a collection of repeated modules.

\begin{figure}[th]
\centering
\includegraphics[width=.2\textwidth]{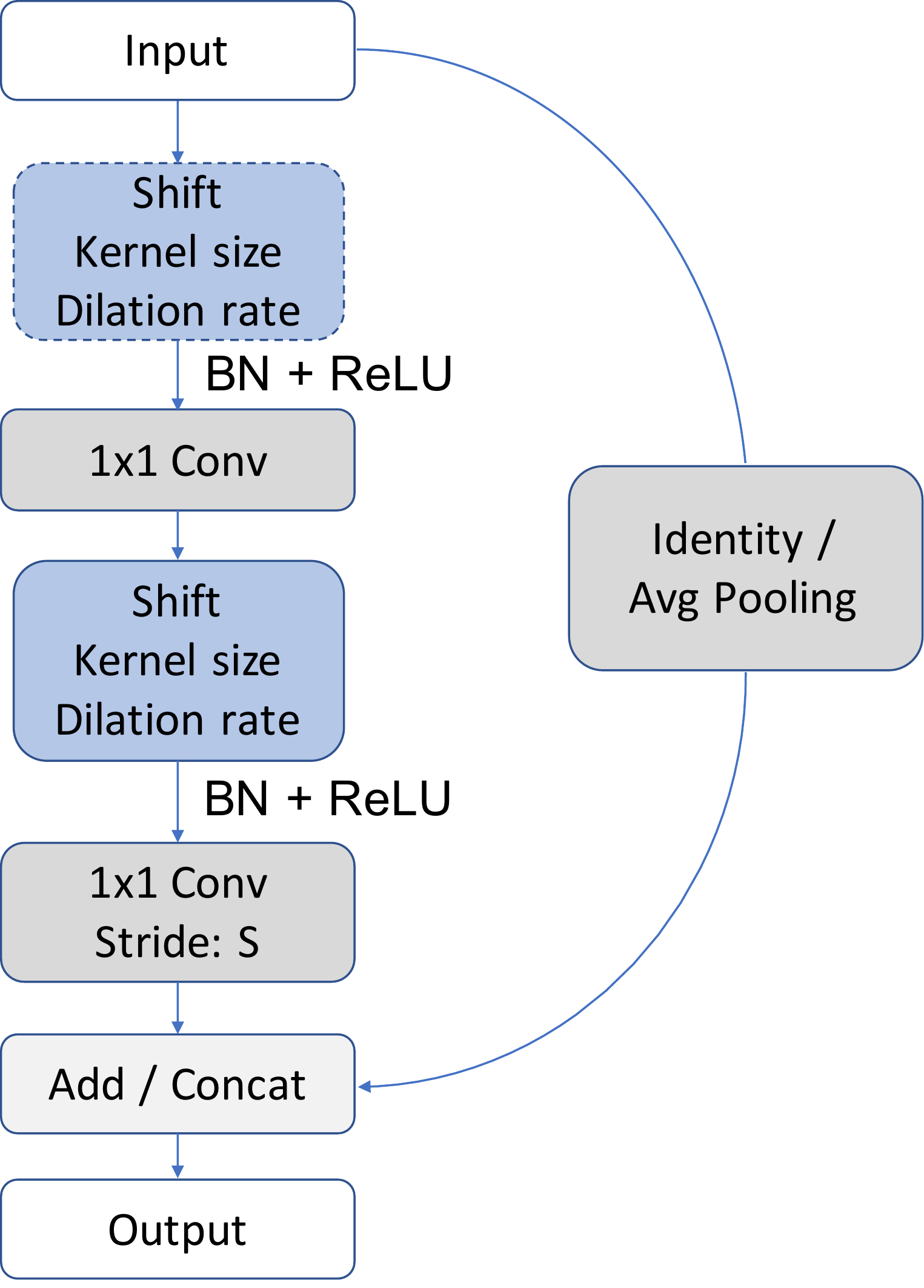}
   \caption{Illustration of the \textit{Conv-Shift-Conv} $CSC$ module and the \textit{Shift-Conv-Shift-Conv} $(SC^2)$ module.}
\label{fig:CSC}
\end{figure}

Based on the analysis in previous sections, we propose a module using shift operations as shown in Figure~\ref{fig:CSC}. The input tensor is first processed by point-wise convolutions. Then, we perform a shift operation to redistribute spatial information. Finally, we apply another set of point-wise convolutions to mix information across channels. Both sets of point-wise convolutions are preceded by batch normalization and a non-linear activation function (ReLU). Following ShuffleNet~\cite{ShuffleNet}, we use an additive residual connection when input and output are of the same shape, and use average pooling with concatenation when we down-sample the input spatially and double the output channels. We refer to this as a \textit{Conv-Shift-Conv} or $CSC$ module. A variant of this module includes another shift operation before the first point-wise convolution; we refer to this as the \textit{Shift-Conv-Shift-Conv} or $SC^2$ module. This allows the designer to further increase the receptive field of the module. 

As with spatial convolutions, shift modules are parameterized by several factors that control its behavior. We use the kernel size of the shift operation to control the receptive field of the $CSC$ module. Akin to the dilated convolution, the ``dilated shift'' samples data at a spatial interval, which we define to be the \textit{dilation rate} $\mathcal{D}$. The \textit{stride} of the $CSC$ module is defined to be the stride of the second point-wise convolution, so that spatial information is mixed in the shift operation before down-sampling. Similar to the bottleneck module used in ResNet, we use the ``expansion rate'' $\mathcal{E}$ to control the intermediate tensor's channel size. With bottleneck modules, 3x3 convolutions in the middle are expensive computationally, forcing small intermediate channel sizes. However, the shift operation allows kernel size $D_F$ adjustments without affecting parameter size and FLOPs. As a consequence, we can employ a shift module to allow larger intermediate channel sizes, where sufficient information can be gathered from nearby positions.

\section{Experiments}

We first assess the shift module's ability to replace convolutional layers, and then adjust hyperparameter $\mathcal{E}$ to observe tradeoffs between model accuracy, model size, and computation. 
We then construct a range of shift-based networks and investigate their performance for a number of different applications.

\subsection{Operation Choice and Hyperparameters}

Using ResNet, we juxtapose the use of 3x3 convolutional layers with the use of $CSC$ modules, by replacing all of ResNet's basic modules (two 3x3 convolutional layers) with $CSC$s to make ``ShiftResNet''. For ResNet and ShiftResNet, we use two Tesla K80 GPUs with batch size 128 and a starting learning rate of 0.1, decaying by a factor of 10 after 32k and 48k iterations, as in \cite{ResNet}. In these experiments, we use the CIFAR10 version of ResNet: a convolutional layer with 16 3x3 filters; 3 groups of basic modules with output channels 16, 32, 64; and a final fully-connected layer. A basic module contains two 3x3 convolutional layers followed by batchnorm and ReLU in parallel with a residual connection. With ShiftResNet, each group contains several $CSC$ modules. We use three ResNet models: in ResNet20, each group contains 3 basic modules. For ResNet56, each contains 5, and for ResNet110, each contains 7. By toggling the hyperparameter $\mathcal{E}$, the number of filters in the $CSC$ module's first set of 1x1 convolutions, we can reduce the number of parameters in ``ShiftResNet'' by nearly 3 times without any loss in accuracy, as shown in Table~\ref{tab:res-shift-vs-conv}. Table~\ref{tab:cifar10-100} summarizes CIFAR10 and CIFAR100 results across all $\mathcal{E}$ and ResNet models.

We next compare different strategies for parameter reduction. We reduce ResNet's parameters to match that of ShiftResNet for some $\mathcal{E}$, denoted ResNet-$\mathcal{E}$ and ShiftResNet-$\mathcal{E}$, respectively. We use two separate approaches: 1) module-wise: decrease the number of filters in each module's first 3x3 convolutional layer; 2) net-wise: decrease every module's input and output channels by the same factor. As Table~\ref{tab:robust} shows, convolutional layers are less resilient to parameter reduction, with the shift module preserving accuracy 8\% better than both reduced ResNet models of the same size, on CIFAR100. In Table~\ref{tab:robust-imagenet}, we likewise find improved resilience on ImageNet as ShiftResNet achieve better accuracy with millions fewer parameters.

\begin{figure}
\begin{tikzpicture}
\begin{axis}[
    title={\textsc{Accuracy vs Parameters Tradeoff}},
    xlabel={Parameters (Millions)},
    ylabel={Accuracy (Top 1 CIFAR100)},
    ylabel style={yshift=-0.5em},
    xmin=0, xmax=2,
    ymin=50, ymax=80,
    xtick={0,0.5,1,1.5,2},
    ytick={50,60,70,80},
    legend pos=south east,
    ymajorgrids=true,
    grid style=dashed,
    width=240pt
]
 
\addplot+[
    purple,
    mark=triangle*,
    solid,
    mark options={solid}
    ]
    coordinates {
    (0.03,55.62)(0.1,62.32)(0.19,68.64)(0.28,69.82)};
\addplot+[
    green!85!black,
    mark=square*,
    solid,
    mark options={solid}
    ]
    coordinates {
    (0.1,65.21)(0.29,69.77)(0.58,72.13)(0.87,73.64)
    };
\addplot[
    mark=*,
    teal!50!cyan
    ]
    coordinates {
    (0.2,67.84)(0.59,71.83)(1.18,72.56)(1.76,74.10)
    };
\addplot+[
    purple,
    mark=diamond*,
    dashed,
    mark options={solid}
    ]
    coordinates {
    (0.03,52.40)(0.1,60.61)(0.19,64.27)(0.28,66.37)};
\addplot+[
    green!85!black,
    mark=triangle*,
    dashed,
    mark options={solid}
    ]
    coordinates {
    (0.1,56.78)(0.29,64.49)(0.58,67.45)(0.87,68.63)
    };
\addplot[mark=square*,teal!50!cyan,dashed]
    coordinates {
    (0.2,60.44)(0.59,66.16)(1.18,68.87)(1.76,70.14)
    };
    \legend{ShiftResNet20,ShiftResNet56,ShiftResNet110,ResNet20,ResNet56,ResNet110}
\end{axis}
\end{tikzpicture}
\caption{This figure shows that ShiftResNet family members are  significantly more efficient than their corresponding ResNet family members. Tradeoff curves further to the top left are more efficient, with higher accuracy per parameter. For ResNet, we take the larger of two accuracies between module-wise and net-wise reduction results.}
\label{fig:acc-v-param-tradeoff}
\end{figure}
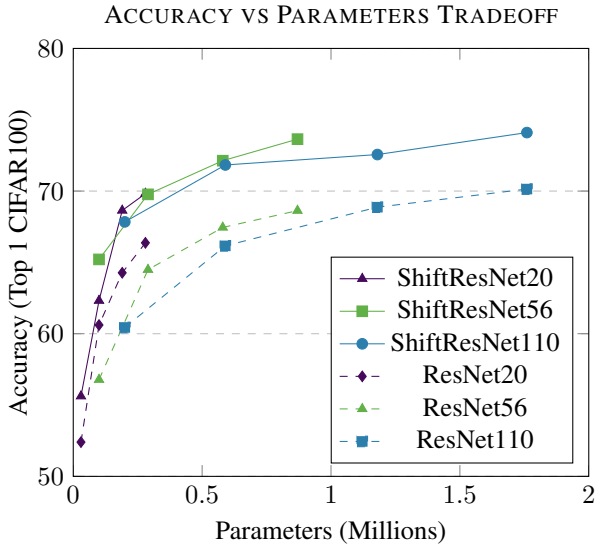

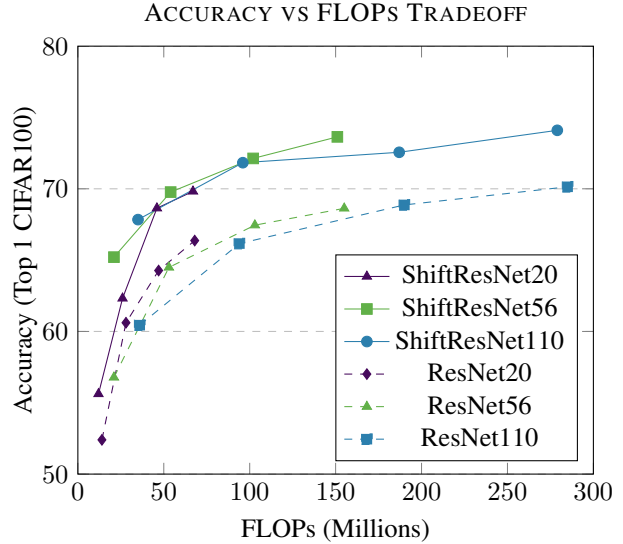
\begin{figure}
\begin{tikzpicture}
\begin{axis}[
    title={\textsc{Accuracy vs FLOPs Tradeoff}},
    xlabel={FLOPs (Millions)},
    ylabel={Accuracy (Top 1 CIFAR100)},
    ylabel style={yshift=-0.5em},
    xmin=0, xmax=300,
    ymin=50, ymax=80,
    xtick={0,50,100,150,200,250,300},
    ytick={50,60,70,80},
    legend pos=south east,
    ymajorgrids=true,
    grid style=dashed,
    width=240pt
]
 
\addplot+[
    purple,
    mark=triangle*,
    solid,
    mark options={solid}
    ]
    coordinates {
    (12,55.62)(26,62.32)(46,68.64)(67,69.82)};
\addplot+[
    green!85!black,
    mark=square*,
    solid,
    mark options={solid}
    ]
    coordinates {
    (21,65.21)(54,69.77)(102,72.13)(151,73.64)
    };
\addplot[mark=*,teal!50!cyan]
    coordinates {
    (35,67.84)(96,71.83)(187,72.56)(279,74.10)
    };
\addplot+[
    purple,
    mark=diamond*,
    dashed,
    mark options={solid}
    ]
    coordinates {
    (14,52.40)(28,60.61)(47,64.27)(68,66.37)};
\addplot+[
    green!85!black,
    mark=triangle*,
    dashed,
    mark options={solid}
    ]
    coordinates {
    (21,56.78)(53,64.49)(103,67.45)(155,68.63)
    };
\addplot[mark=square*,
    teal!50!cyan,
    dashed]
    coordinates {
    (36,60.44)(94,66.16)(190,68.87)(285,70.14)
    };
    \legend{ShiftResNet20,ShiftResNet56,ShiftResNet110,ResNet20,ResNet56,ResNet110}
\end{axis}
\end{tikzpicture}
\caption{Tradeoff curves further to the top left are more efficient, with higher accuracy per FLOP. This figure shows ShiftResNet is more efficient than ResNet, in FLOPs.}
\label{fig:acc-v-flops-tradeoff}
\end{figure}

Table~\ref{tab:shrink-model-size} shows that ShiftResNet consistently outperforms ResNet, when both are constrained to use 1.5x fewer parameters. Table~\ref{tab:cifar10-100} then includes all results. Figure~\ref{fig:acc-v-param-tradeoff} shows the tradeoff between CIFAR100 accuracy and number of parameters for the hyperparameter $\mathcal{E} \in \{1,3,6,9\}$ across both \{ResNet, ShiftResNet\} models using varying numbers of layers $\ell \in \{20,56,110\}$. Figure~\ref{fig:acc-v-flops-tradeoff} examines the same set of possible models and hyperparameters but between CIFAR100 accuracy and the number of FLOPs. Both figures show that ShiftResNet models provide superior trade-off between accuracy and parameters/FLOPs. 

\begin{table}[h]
\begin{center}
\caption{\textsc{Parameters for Shift vs Convolution, with Fixed Accuracy on CIFAR-100}}
\label{tab:res-shift-vs-conv}
\vspace{-0.1in}
\begin{tabular}{c | c c c c}
Model & \thead{Top1 Acc} & \thead{FLOPs} & \thead{Params} \\
\hline
ShiftResNet56-3 & \textbf{69.77}\% & \textbf{54M} & \textbf{0.29M}\\
ResNet56 & 69.27\% & 151M  & 0.87M \\
\end{tabular}
\end{center}
\vspace{-0.2in}
\end{table}

\begin{table}
\begin{center}
\caption{\textsc{Reduction Resilience for Shift vs Convolution, with Fixed Parameters}}
\label{tab:robust}
\vspace{-0.1in}
\begin{tabular}{c | c c c c}
Model & \thead{CIFAR-100 Acc} & \thead{FLOPs} & \thead{Params} \\
\hline
ShiftResNet110-1 & \textbf{67.84}\% & 36M & \textbf{203K} \\
ResNet110-1 & 60.44\% & \textbf{32M} & 211K \\
\end{tabular}
\end{center}
\vspace{-0.2in}
\end{table}

\begin{table}
\begin{center}
\caption{\textsc{Reduction Resilience for Shift vs \\Convolution on ImageNet}}
\label{tab:robust-imagenet}
\vspace{-0.1in}
\begin{tabular}{c c c}
\thead{ {\normalsize Shift50} \\ Top1 / Top5 Acc} & \thead{{\normalsize ResNet50} \\ Top1 / Top5 Acc} & \thead{{\normalsize Parameters} \\ Shift50 / ResNet50} \\
\hline
\textbf{75.6} / \textbf{92.8} & 75.1 / 92.5 & 22M / 26M\\
\textbf{73.7} / \textbf{91.8} & 73.2 / 91.6 & 11M / 13M \\
\textbf{70.6} / 89.9 & 70.1 / 89.9 & 6.0M / 6.9M \\
\hline
\end{tabular}
\caption*{Here, we abbreviate ``ShiftResNet50'' as ``Shift50''.}
\end{center}
\vspace{-0.2in}
\end{table}

\begin{table}
\begin{center}
\caption{\textsc{Performance Across ResNet 
Models, with 1.5 Fewer Parameters}}
\label{tab:shrink-model-size}
\vspace{-0.1in}
\begin{tabular}{c | c c c c}
No. Layers & \thead{ {\normalsize ShiftResNet-6} \\ CIFAR100 Top 1} & \thead{{\normalsize ResNet } \\ CIFAR100 Top 1} \\
\hline
20 & \textbf{68.64\%} & 66.25\%\\
56 & \textbf{72.13\%} & 69.27\%\\
110 & \textbf{72.56\%} & 72.11\%\\
\end{tabular}
\end{center}
\vspace{-0.2in}
\end{table}

\begin{table*}
\caption{\textsc{Shift Operation Analysis Using CIFAR10 and CIFAR100}}
\label{tab:cifar10-100}
\vspace{-0.1in}
\begin{tabular}{c | c c c c c c c c}
Model & $\mathcal{E}$ & \thead{ {\normalsize ShiftResNet} \\ CIFAR10 / 100 Accuracy} & \thead{{\normalsize ResNet (Module) }\\ CIFAR10 / 100 Accuracy} &
\thead{{\normalsize ResNet (Net)} \\ CIFAR100 Accuracy} &\thead{{ \normalsize ShiftResNet }\\ Params / FLOPs ($\times10^6$)} & \thead{{\normalsize Reduction Rate} \\ Params / FLOPs}\\
\hline
20 & 1 & 86.66\% / 55.62\% & 85.54\% / 52.40\% & 49.58\% & 0.03 / 12 & 7.8 / 5.5\\
20 & 3 & 90.08\% / 62.32\% & 88.33\% / 60.61\% & 58.16\% & 0.10 / 26 & 2.9 / 2.6\\
20 & 6 & 90.59\% / 68.64\% & 90.09\% / 64.27\% & 63.22\% & 0.19 / 46 & 1.5 / 1.4 \\
20 & 9 & \textbf{91.69\%} / \textbf{69.82\%} & 91.35\% / 66.25\% & 66.25\% & 0.28 / 67 & 0.98 / 1.0 \\
\hline
56 & 1 & 89.71\% / 65.21\% & 87.46\% / 56.78\% & 56.62\% & 0.10 / 21 & 8.4 / 7.0\\
56 & 3 & 92.11\% / 69.77\% & 89.40\% / 62.53\% & 64.49\% & 0.29 / 54 & 2.9 / 2.8\\
56 & 6 & 92.69\% / 72.13\% & 89.89\% / 61.99\% & 67.45\% & 0.58 / 102 & 1.5 / 1.5 \\
56 & 9 & \textbf{92.74\%} / \textbf{73.64\%} & 92.01\% / 69.27\% & 69.27\% & 0.87 / 151 & 0.98 / 1.0 \\
\hline
110 & 1 & 90.34\% / 67.84\% & 76.82\% / 39.90\% & 60.44\% & 0.20 / 35 & 8.5 / 7.8\\
110 & 3 & 91.98\% / 71.83\% & 74.30\% / 40.52\% & 66.61\% & 0.59 / 96 & 2.9 / 2.9\\
110 & 6 & 93.17\% / 72.56\% & 79.02\% / 40.23\% & 68.87\% & 1.18 / 187 & 1.5 / 1.5\\
110 & 9 & \textbf{92.79\%} / \textbf{74.10\%} & 92.46\% / 72.11\% & 72.11\% & 1.76 / 279 & 0.98 / 1.0\\
\hline
\end{tabular}
\caption*{Note that the ResNet-9 results are replaced with accuracy of the original model. The number of parameters holds for both ResNet and ShiftResNet across CIFAR10, CIFAR100. FLOPs are computed for ShiftResNet. All accuracies are Top 1. ``Reduction Rate'' is the original ResNet's parameters/flops over the new ShiftResNet's parameters/flops.}
\vspace{-0.3in}
\end{table*}

\subsection{ShiftNet}

Even though ImageNet classification is not our primary goal, to further investigate the effectiveness of the shift operation, we use the proposed $CSC$ module shown in Figure~\ref{fig:CSC} to design a class of efficient models called ShiftNet and present its classification performance on standard benchmarks to compare with state-of-the-art small models. 

Since an external memory access consumes 1000x more energy than a single arithmetic operation \cite{effcient_learning}, our primary goal in designing ShiftNet is to optimize the number of parameters and thereby to reduce memory footprint.
In addition to the general desirability of energy efficiency, 
the main targets of ShiftNet are mobile and IOT applications, 
where memory footprint, even
more so than FLOPs, are a primary constraint.
In these application domains small models can be packaged in mobile and IOT applications that are delivered within the 100-150MB limit for mobile over-the-air updates. In short, our design goal for ShiftNets is to attain competitive accuracy with fewer parameters.

The ShiftNet architecture is described in Table~\ref{tab:shiftnet-arch}. Since parameter size does not grow with shift kernel size, we use a larger kernel size of 5 in earlier modules. We adjust the expansion parameter $\mathcal{E}$ to scale the parameter size in each $CSC$ module. We refer to the architecture described in Table~\ref{tab:shiftnet-arch} as ShiftNet-A. We shrink the number of channels in all $CSC$ modules by 2 for ShiftNet-B. We then build a smaller and shallower network, with $\{1, 4, 4, 3\}$ CSC modules in groups $\{1, 2, 3, 4\}$ with channel sizes of $\{32, 64, 128, 256\}$, $\mathcal{E}=1$ and kernel size is 3 for all modules. We name this shallow model ShiftNet-C. 
We train the three ShiftNet variants on the ImageNet 2012 classification dataset~\cite{ILSVRC15} with 1.28 million images and evaluate on the validation set of 50K images. We adopt data augmentations suggested by ~\cite{Xception} and weight initializations suggested by~\cite{ResNet}. We train our models for 90 epochs on 64 Intel KNL instances using Intel Caffe~\cite{caffe} with a batch size of 2048, an initial learning rate of 0.8, and learning rate decay by $10$ every 30 epochs. 

In Table~\ref{tab:shiftnet_result}, we show classification accuracy and number of parameters for ShiftNet and other state-of-the-art models. We compare ShiftNet-\{A, B, C\} with 3 groups of models with similar levels of accuracy. In the first group, ShiftNet-A is \textbf{34X} smaller than VGG-16, while the top-1 accuracy drop is only $1.4\%$. ShiftNet-B's top-1 accuracy is $2.5\%$ worse than its MobileNet counterpart, but it uses fewer parameters. We compare ShiftNet-C with SqueezeNet and AlexNet, and we can achieve better accuracy with $2/3$ the number of SqueezeNet's parameters, and \textbf{77X} smaller than AlexNet.

\begin{table}[]
\centering
\caption{\textsc{ShiftNet Architecture}}
\label{tab:shiftnet-arch}
\begin{tabular}{c|c|c|c|c|c}
\hline\hline
Group & \begin{tabular}[c]{@{}c@{}}Type/\\ Stride\end{tabular} & Kernel            & $\mathcal{E}$ & \begin{tabular}[c]{@{}c@{}}Output\\ Channel\end{tabular} & Repeat \\ \hline
- & Conv /s2      & 7$\times$7      &   -     & 32                                                       & 1      \\ \hline 
1 & $\mbox{CSC}$ / s2      & 5$\times$5      & 4      & 64                                                       & 1      \\ 
\newline & $\mbox{CSC}$ / s1      & 5$\times$5      & 4      &                                                        & 4      \\ \hline
2 & $\mbox{CSC}$ / s2      & 5$\times$5      & 4      & 128                                                      & 1      \\ 
\newline & $\mbox{CSC}$ / s1      & 5$\times$5      & 3      &                                                       & 5      \\ \hline 
3 & $\mbox{CSC}$ / s2      & 3$\times$3      & 3      & 256                                                      & 1      \\ 
\newline & $\mbox{CSC}$ / s1      & 3$\times$3      & 2      &                                                      & 6      \\ \hline 
4 & $\mbox{CSC}$ / s2      & 3$\times$3      & 2      & 512                                                      & 1      \\ 
\newline & $\mbox{CSC}$ / s1      & 3$\times$3      & 1      &                                                       & 2      \\ \hline 
- & Avg Pool      & 7$\times$7  &   -     & 512                                                      & 1      \\ \hline
- & FC            & - &     -     & 1k                                                       & 1      \\ \hline
\end{tabular}
\vspace{-0.1in}
\end{table}

\begin{table}[]
\centering
\caption{\textsc{ShiftNet Results on Imagenet}}
\label{tab:shiftnet_result}
\begin{tabular}{c|cc}
\hline\hline
Model  & \thead{Accuracy\\ Top-1 / Top-5} & \thead{Parameters \\ (Millions)} \\ \hline
VGG-16~\cite{VGG} & \textbf{71.5} / \textbf{90.1} & 138 \\
GoogleNet~\cite{GoogleNet} & 69.8 / - & 6.8 \\
ShiftResNet-0.25 (ours) & 70.6 / 89.9 & 6.0 \\
ShuffleNet-2$\times$~\cite{ShuffleNet}* & 70.9 / - & 5.6 \\
1.0 MobileNet-224~\cite{MobileNet} & 70.6 / - & 4.2\\
Compact DNN~\cite{compactDNN} & 68.9 / 89.0 & 4.1 \\
ShiftNet-A (ours)  & 70.1 / 89.7 & \textbf{4.1} \\
\hline
0.5 MobileNet-224~\cite{MobileNet} & \textbf{63.7} / - & 1.3\\ 
ShiftNet-B (ours) & 61.2 / 83.6 & \textbf{1.1} \\
\hline
AlexNet~\cite{AlexNet} & 57.2 / 80.3 & 60 \\
SqueezeNet~\cite{SqueezeNet} & 57.5 / 80.3 & 1.2 \\
ShiftNet-C (ours) & \textbf{58.8} / \textbf{82.0} & \textbf{0.78} \\
\hline 
\end{tabular}
    \caption*{* Estimated according to the model description in~\cite{ShuffleNet}}
\vspace{-0.2in}
\end{table}

\subsection{Face Embedding}

We continue to investigate the shift operation for different applications. Face verification and recognition are becoming increasingly popular on mobile devices. Both functionalities rely on face embedding, which aims to learn a mapping from face images to a compact embedding in Euclidean space, where face similarity can be directly measured by embedding distances. Once the space has been generated, various face-learning tasks, such as facial recognition and verification, can be easily accomplished by standard machine learning methods with feature embedding. Mobile devices have limited computation resources, therefore creating small neural networks for face embedding is a necessary step for mobile deployment.

FaceNet~\cite{schroff2015facenet} is one state-of-the-art face embedding approach. The original FaceNet is based on Inception-Resnet-v1~\cite{szegedy2016rethinking}, which contains 28.5 million parameters, making it difficult to be deployed on mobile devices. We propose a new model ShiftFaceNet based on ShiftNet-C from the previous section, which only contains 0.78 million parameters. 

Following~\cite{parkhi2015deep}, we train FaceNet and ShiftFaceNet by combining the softmax loss with center loss~\cite{wen2016discriminative}. We evaluate the proposed method on three datasets for face verification: given a pair of face images, a distance threshold is selected to classify the two images belonging to the same or different entities. The LFW dataset~\cite{huang2007labeled} consists of 13,323 web photos with 6,000 face pairs.
The YTF dataset~\cite{wolf2011face} includes 3,425 with 5,000 video pairs for video-level face verification.
The MS-Celeb-1M dataset (MSC)~\cite{guo2016ms} comprises 8,456,240 images for 99,892 entities.
In our experiments, we randomly select 10,000 entities from MSC as our training set, to learn the embedding space. We test on 6,000 pairs from LFW, 5,000 pairs from YTF and 100,000 pairs randomly generated from MSC, excluding the training set.
In the pre-processing step, we detect and align all faces using a multi-task CNN~\cite{zhang2016joint}.
Following~\cite{schroff2015facenet},
the similarity between two videos is computed as the average similarity of 100 random pairs of frames, one from each video. Results are shown in Table~\ref{tab:face_accuracy}. Parameter size for the original FaceNet and our proposed ShiftFaceNet is shown in Table ~\ref{tab:face_parameter}. With ShiftFaceNet, we are able to reduce the parameter size by \textbf{35X}, with at most $2\%$ drop of accuracy in above three verification benchmarks. 

\begin{table}
\small
\caption{\textsc{Face Verification Accuracy for ShiftFaceNet vs FaceNet~\cite{schroff2015facenet}.}}
\vspace{-0.1in}
\label{tab:face_accuracy}
\begin{tabular}{c | c c | c c}
 & \multicolumn{2}{c|}{Accuracy$\pm$ STD (\%)} & \multicolumn{2}{c}{Area under curve (\%)} \\
\hline
 & FaceNet & ShiftFaceNet & FaceNet & ShiftFaceNet \\
LFW & 97.1$\pm$1.3 & 96.0$\pm$1.4 & 99.5 & 99.4\\
YTF & 92.0$\pm$1.1 & 90.1$\pm$0.9 & 97.3 & 96.1\\
MSC & 79.2$\pm$1.7 & 77.6$\pm$1.7 & 85.6 & 84.4\\
\end{tabular}
\end{table}

\begin{table}
\begin{center}
\caption{\textsc{Face Verification Parameters for ShiftFaceNet vs FaceNet~\cite{schroff2015facenet}.}}
\vspace{-0.1in}
\label{tab:face_parameter}
\begin{tabular}{c | c c}
Model & FaceNet & ShiftFaceNet \\
\hline
Params (Millions) & 28.5 & 0.78\\
\end{tabular}
\end{center}
\vspace{-.3in}
\end{table}

\subsection{Style Transfer}
Artistic style transfer is another popular application on mobile devices. It is an image transformation task where the goal is to combine the \textit{content} of one image with the \textit{style} of another. Although this is an ill-posed problem without definite quantitative metrics, a successful style transfer requires that networks capture both minute textures and holistic semantics, for content and style images.

Following \cite{gatys-style-transfer, johnson-style-transfer}, we use perceptual loss functions to train a style transformer. In our experiment, we use a VGG-16 network pretrained on ImageNet to generate the perceptual loss. The original network trained by Johnson \textit{et al.} \cite{johnson-style-transfer} consists of three downsampling convolutional layers, five residual modules, and three upsampling convolutional layers. All non-residual convolutions are followed by an instance normalization layer \cite{instance-norm}. In our experiments, we replace all but the first and last convolution layers with shifts followed by 1x1 convolutions. We train the network on the COCO \cite{coco} dataset using the previously reported hyperparameter settings $\lambda_s \in \{1e10, 5e10, 1e11\}, \lambda_c = 1e5$. By replacing convolutions with shifts, we achieve an overall \textbf{6X} reduction in the number of parameters with minimal degredation in image quality. Examples of stylized images generated by original and shift based transformer networks can be found in Figure~\ref{fig:style}.

\newcommand{\STincludegraphics}[2][]{\includegraphics[width = 0.9in, height=0.9in,#1]{#2}}
{
\setlength\tabcolsep{2pt}
\begin{figure*}
\begin{tabular}{c | c c c | c c c}
\textsc{Content} & \textsc{Style} &  \textsc{Shift}&  \textsc{Original} & \textsc{Style} & \textsc{Shift} &  \textsc{Original} \\
\subfloat{\STincludegraphics[]{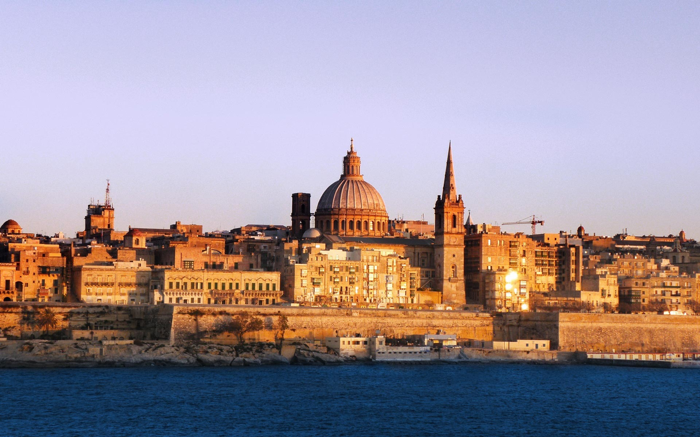}} &
 \multirow{1}{*}[4ex]{\subfloat{\STincludegraphics[]{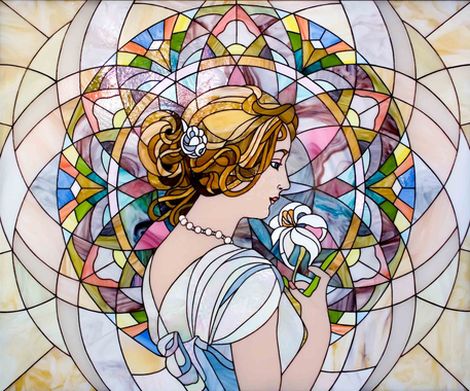}}} &
\subfloat{\STincludegraphics[]{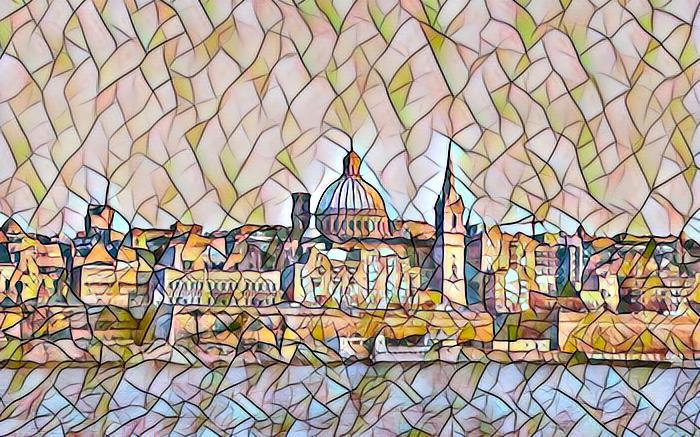}} &
\subfloat{\STincludegraphics[]{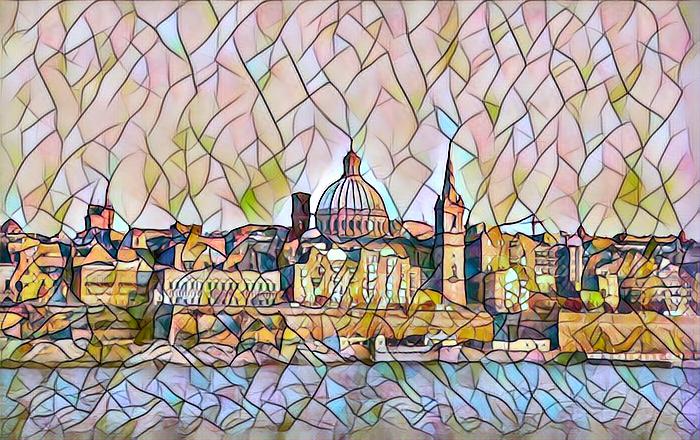}} &
\multirow{1}{*}[4ex]{\subfloat{\STincludegraphics[]{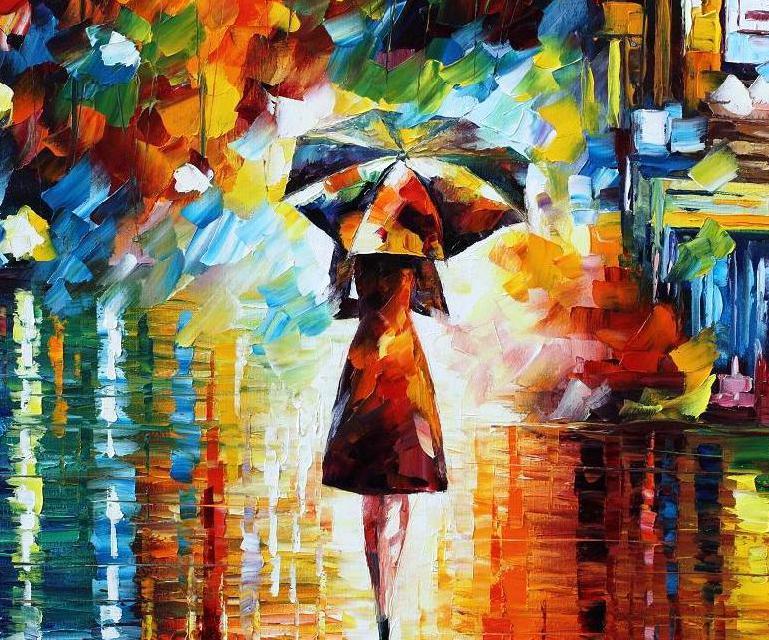}}} &
\subfloat{\STincludegraphics[]{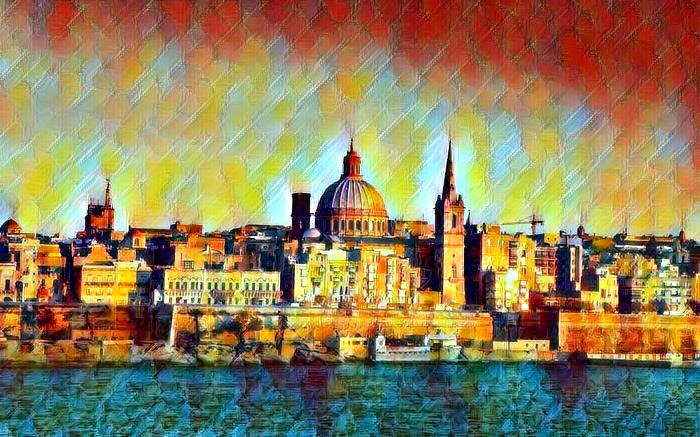}} &
\subfloat{\STincludegraphics[]{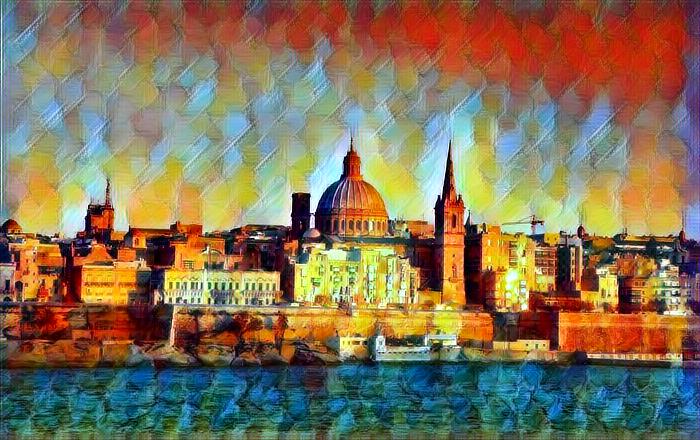}} \\
\subfloat{\STincludegraphics[]{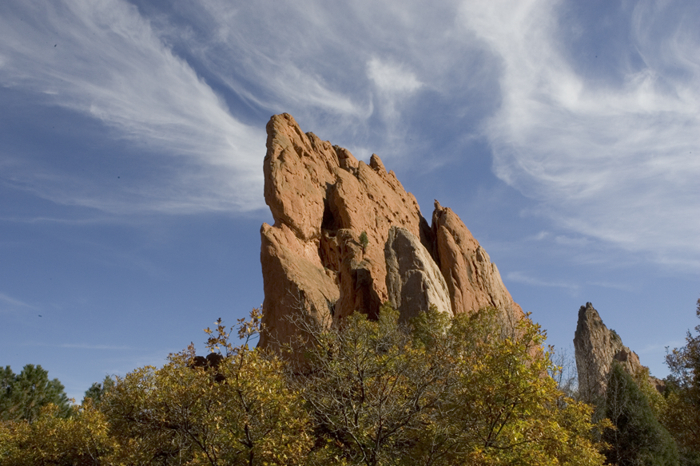}} & 
&
\subfloat{\STincludegraphics[]{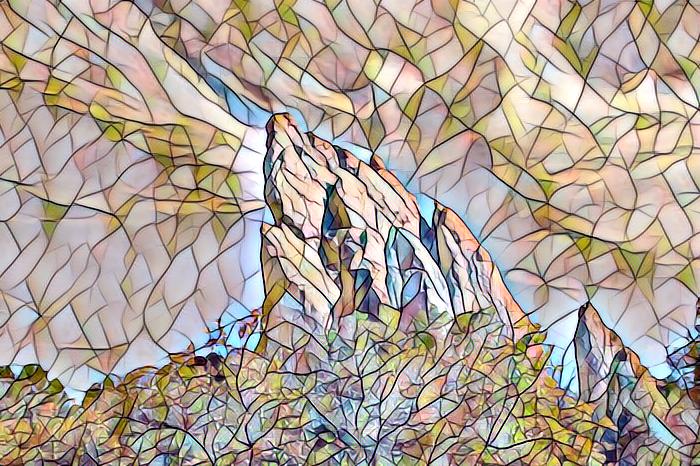}} &
\subfloat{\STincludegraphics[]{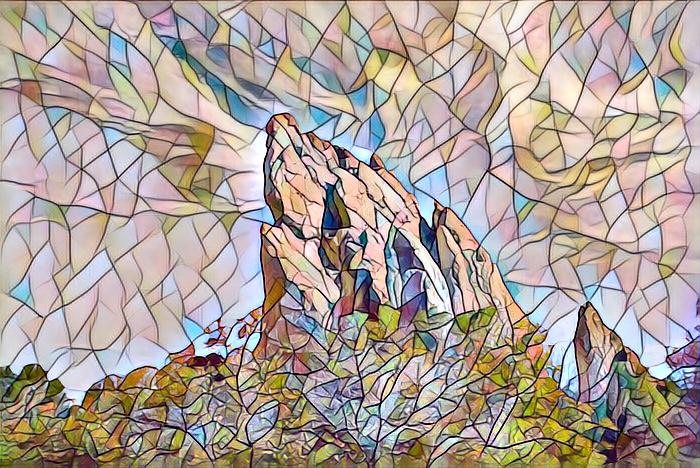}} &
&
\subfloat{\STincludegraphics[]{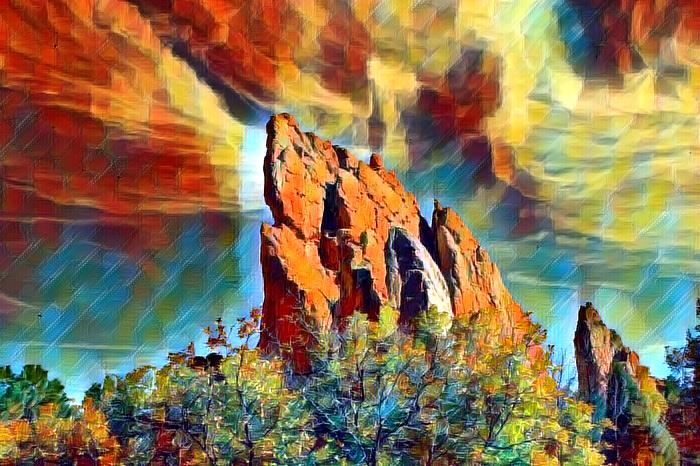}} & 
\subfloat{\STincludegraphics[]{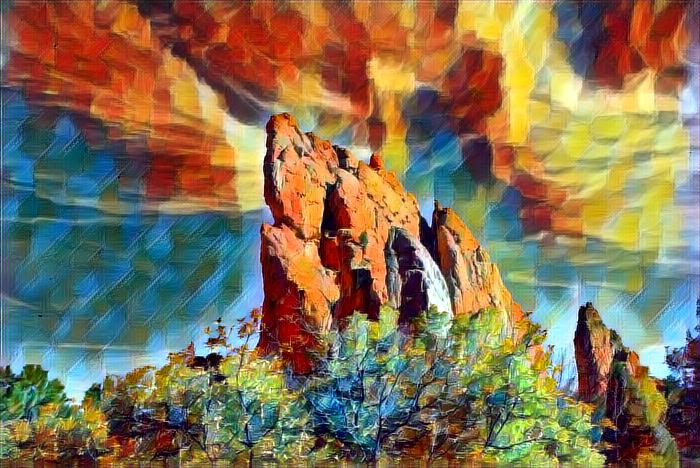}} 
\end{tabular}
\caption{\textsc{Style Transfer Results Using ShiftNet}}
\label{fig:style}
\end{figure*}
}

\begin{table}
\begin{center}
\caption{\textsc{Style Transfer: Shift vs Convolution}}
\vspace{-0.1in}
\begin{tabular}{c | c c }
Model & Original & Shift \\
\hline
Params (Millions)  & 1.9   & 0.3 
\end{tabular}
\end{center}
\vspace{-.3in}
\end{table}

\begin{figure*}
\begin{tikzpicture}
  \begin{axis}[
    colorbar,
    colorbar style={
        major tick length=0pt
    },
    point meta min=0, point meta max=1,
    scatter,
    scatter src=y,
    only marks,
    clip mode=individual,
    scatter/@pre marker code/.append code={
            \pgfkeys{/pgf/fpu=true,/pgf/fpu/output format=fixed}
            \pgfmathsetmacro\negheight{-\pgfplotspointmeta}         
            \fill [draw=black] (axis direction cs:-0.5,0) rectangle (axis direction cs:0.5,\negheight);
            \pgfplotsset{mark=none}
        },
    width=12cm,
    height=4.5cm,
    ymajorgrids=true,
    ytick={0,0.2,0.4,0.6,0.8,1},
    xtick={8,24,40,56,72,86,104,120,136},
    xticklabels={TL,TM,TR,ML,MM,MR,BL,BM,BR},
    xmin=-0.5,
    xmax=143.5,
    xlabel={Shift},
    ylabel={Contribution},
    ymin=0,
    ymax=1,
  ]![enter image description here][3]
    \addplot coordinates
      {(0,0.10) (1,0.18) (2,0.21) (3,0.24) (4,0.27) (5,0.27) (6,0.29) (7,0.29) (8,0.32) (9,0.33) (10,0.34) (11,0.34) (12,0.40) (13,0.41) (14,0.43) (15,0.48) (16,0.00) (17,0.00) (18,0.01) (19,0.17) (20,0.19) (21,0.28) (22,0.31) (23,0.33) (24,0.35) (25,0.36) (26,0.39) (27,0.45) (28,0.47) (29,0.67) (30,0.81) (31,0.94) (32,0.02) (33,0.10) (34,0.15) (35,0.16) (36,0.18) (37,0.19) (38,0.24) (39,0.24) (40,0.25) (41,0.26) (42,0.27) (43,0.31) (44,0.34) (45,0.38) (46,0.41) (47,0.41) (48,0.00) (49,0.11) (50,0.20) (51,0.26) (52,0.33) (53,0.34) (54,0.37) (55,0.37) (56,0.40) (57,0.42) (58,0.46) (59,0.53) (60,0.54) (61,0.68) (62,0.97) (63,1.00) (64,0.00) (65,0.11) (66,0.20) (67,0.23) (68,0.24) (69,0.27) (70,0.27) (71,0.30) (72,0.38) (73,0.41) (74,0.45) (75,0.46) (76,0.46) (77,0.65) (78,0.68) (79,0.83) (80,0.21) (81,0.23) (82,0.25) (83,0.26) (84,0.32) (85,0.37) (86,0.42) (87,0.48) (88,0.49) (89,0.49) (90,0.58) (91,0.60) (92,0.69) (93,0.74) (94,0.78) (95,0.80) (96,0.00) (97,0.02) (98,0.14) (99,0.15) (100,0.17) (101,0.21) (102,0.24) (103,0.29) (104,0.30) (105,0.30) (106,0.30) (107,0.31) (108,0.37) (109,0.38) (110,0.41) (111,0.49) (112,0.00) (113,0.06) (114,0.09) (115,0.16) (116,0.18) (117,0.21) (118,0.22) (119,0.24) (120,0.25) (121,0.27) (122,0.28) (123,0.32) (124,0.35) (125,0.62) (126,0.77) (127,0.78) (128,0.11) (129,0.15) (130,0.17) (131,0.19) (132,0.20) (133,0.21) (134,0.22) (135,0.30) (136,0.36) (137,0.37) (138,0.45) (139,0.45) (140,0.46) (141,0.49) (142,0.55) (143,0.58)};
  \end{axis}
\end{tikzpicture}
\begin{tikzpicture}
\begin{axis}[
  width=4.5cm,
  height=4.5cm,
  scatter/@pre marker code/.append code={
            \pgfkeys{/pgf/fpu=true,/pgf/fpu/output format=fixed}
            \fill [draw=black] (axis direction cs:0,0) rectangle (axis direction cs:1,1);
            \pgfplotsset{mark=none}
        },
  xmin=0, xmax=3, ymin=0, ymax=3,
  point meta min=0, point meta max=1,
  xtick={0.5,1.5,2.5},
  xticklabels={left,mid,right},
  ytick={0.5,1.5,2.5},
  yticklabels={bot,mid,top},
  xlabel={\color{white}{Easter egg 1!}}
  ],
  \addplot[only marks, scatter, scatter src=explicit,
  mark size=10]
  coordinates {
  (0,2) [.26]
  (1,2) [.48]
  (2,2) [.0]

  (0,1) [.81]
  (1,1) [.53]
  (2,1) [1.00]

  (0,0) [.04]
  (1,0) [.23]
  (2,0) [.36]
  };
\end{axis}
\end{tikzpicture}
\vspace{-.1in}
\caption*{\hspace{.8in} Normalized Channel Contributions for ResNet20 \hspace*{1.6in} Per-Shift Contributions}
\caption{Left: We plot normalized channel contributions. Along the horizontal axis, we rank the channels in 9 groups of 16 (one group per shift) B=bottom, M=mid, T=top, R=right, L=left. Right: For each shift pattern, we plot the sum of its contributions normalized to the same scale. Both figures share the same color map, where yellow has the highest magnitude.}
\label{fig:NCC}
\vspace{-.2in}
\end{figure*}

\section{Discussion}
In our experiments, we demonstrate the shift operation's effectiveness as an alternative to spatial convolutions. Our construction of shift groups is trivial: we assign a fixed number of channels to each shift. However, this assignment is largely uninformed. In this section, we explore more informed allocations and potential improvements.

An ideal channel allocation should at least have the following 2 properties: 1) Features in the same shift group should not be redundant. We can measure redundancy by checking the correlation between channel activations within a shift group. 2) Each shifted feature should have a non-trivial contribution to the output. We can measure the the contribution of channel-$m$ to the output by $\|P_{m, :}\|_2$, $i.e.$, the $l2$ norm of the $m$-th row of the second point-wise convolution kernel in a $CSC$ module. 

\subsection{Channel Correlation Within Shifts}
We analyze one $CSC$ module with 16 input/output channels and an expansion of 9, from a trained ShiftResnet20 model. We first group channels by shift. Say each shift group contains $c$ channels. We consider activations of the $c$ channels as a random variable $X \in \mathbb{R}^c$. We feed validation images from CIFAR100 into the network and record intermediate activations in the $CSC$ module, to estimate correlation matrices $\Sigma_{XX} \in \mathbb{R}^{c\times c}$, a few of which are shown in figure~\ref{fig:correlation}. We use correlation between channels as a proxy for measuring redundancy. For example, if the correlation between two channels is above a threshold, we can remove one channel. In Figure~\ref{fig:correlation}, candidate pairs could stem from the bright green pixels slightly off-center in the first matrix.

\newcommand{\CVincludegraphics}[2][]{\includegraphics[trim={3cm 0.75cm 3cm 1.25cm},clip,width=0.75in,height=0.75in,#1]{#2}}
{
\setlength\tabcolsep{2pt}
\begin{figure}
\label{tab:cov}
\begin{center}
\begin{tabular}{c c c c}
\subfloat{\CVincludegraphics[]{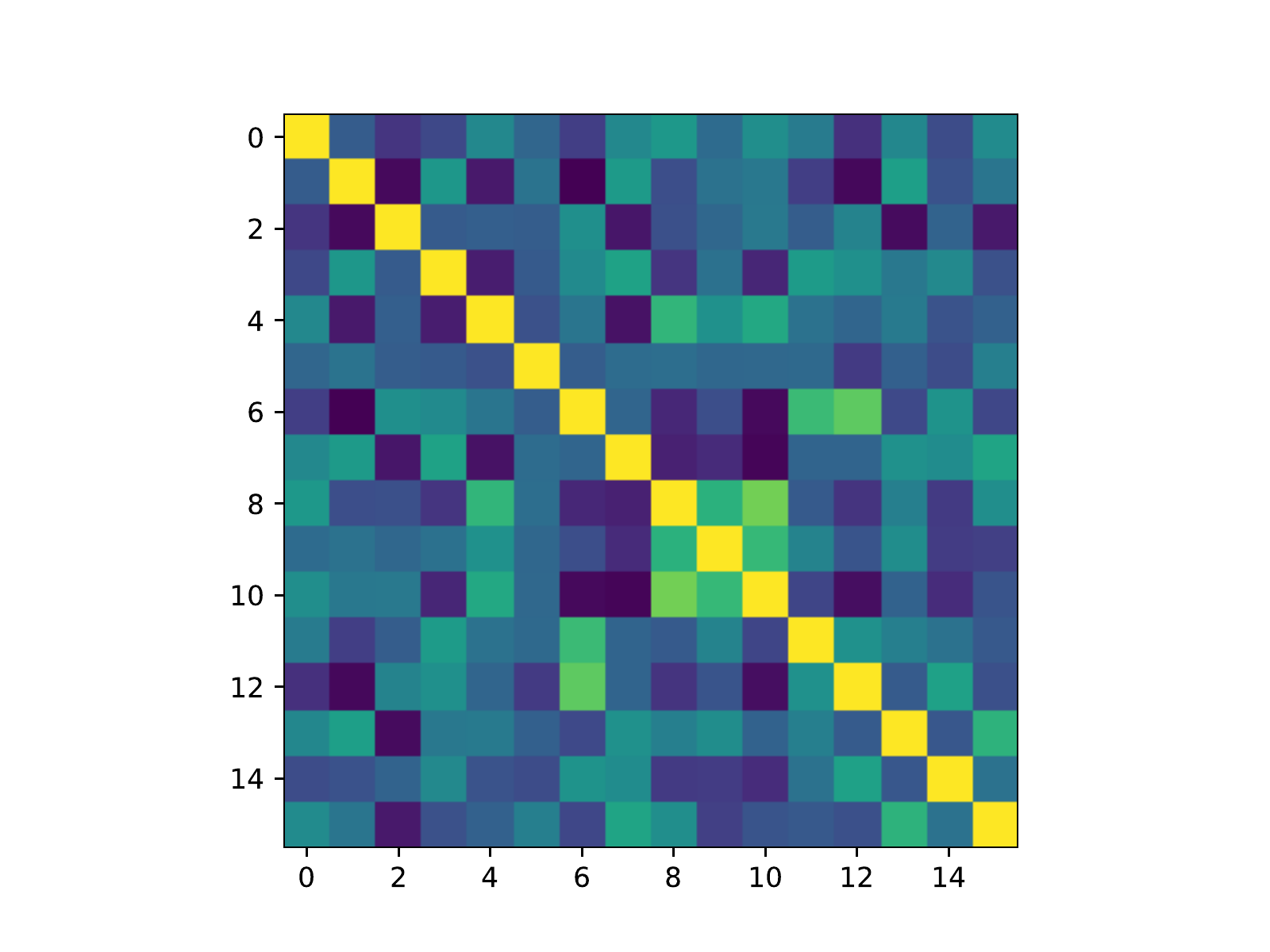}} &
\subfloat{\CVincludegraphics[]{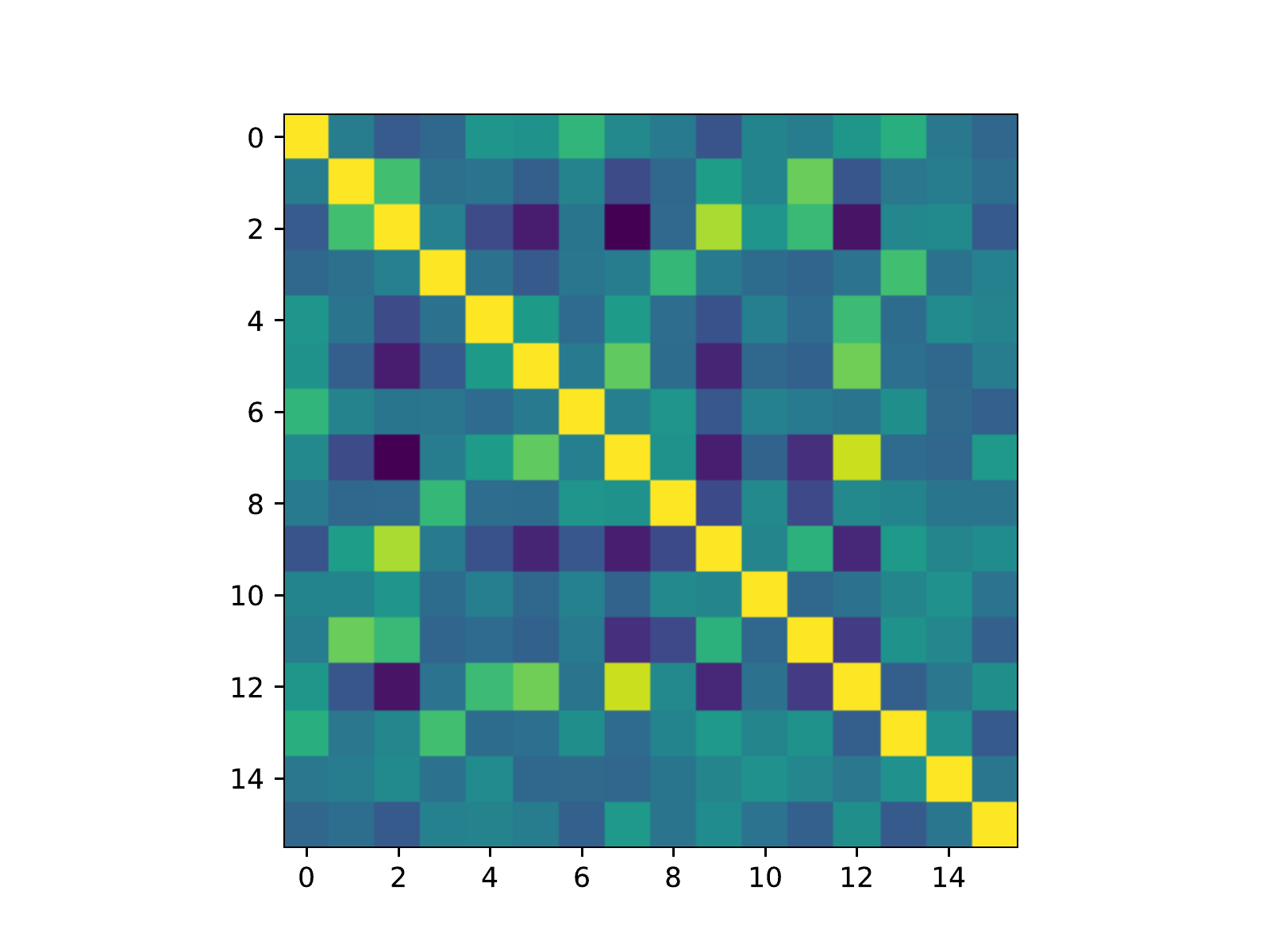}} &
\subfloat{\CVincludegraphics[]{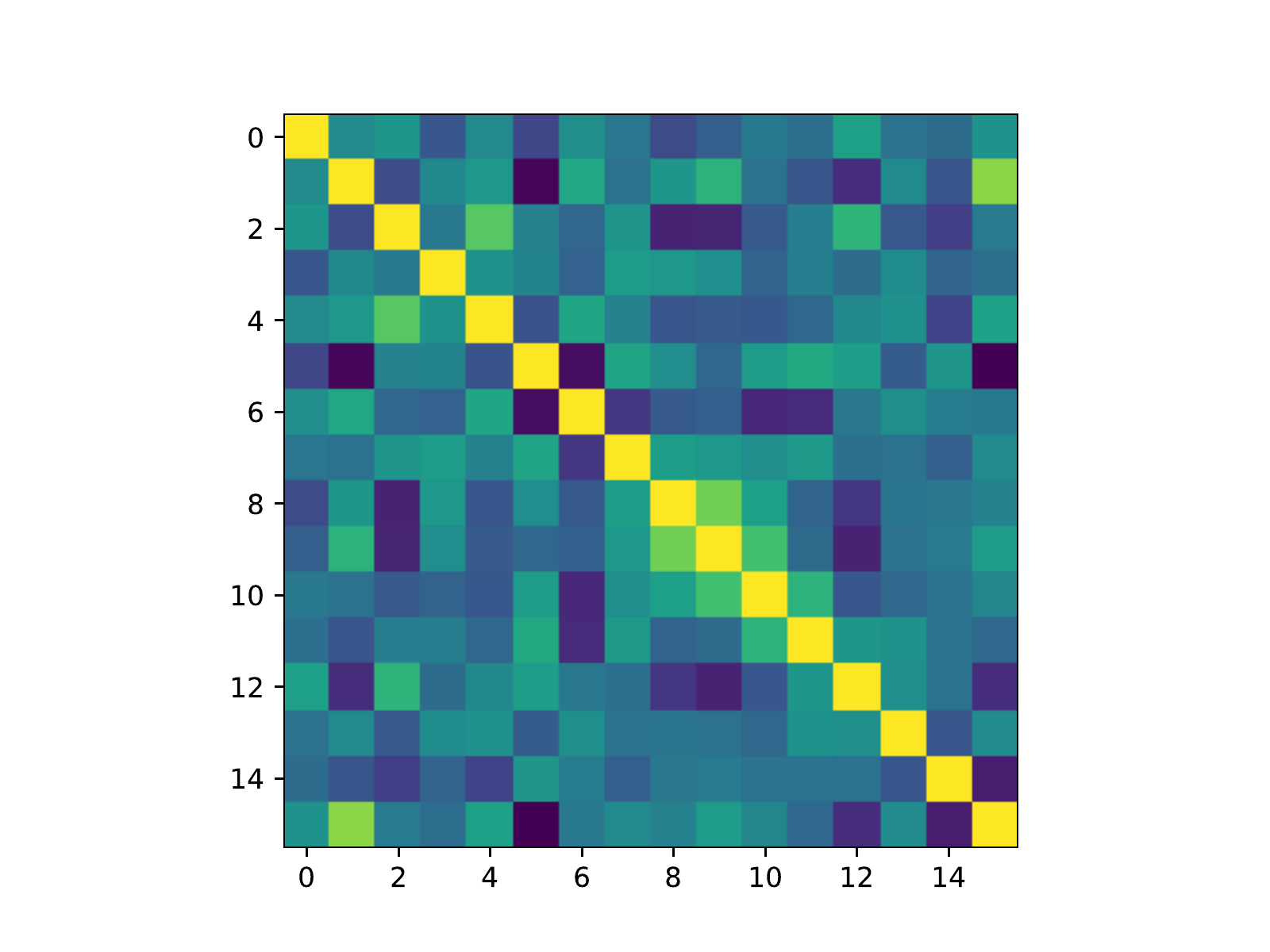}} &
\subfloat{\CVincludegraphics[]{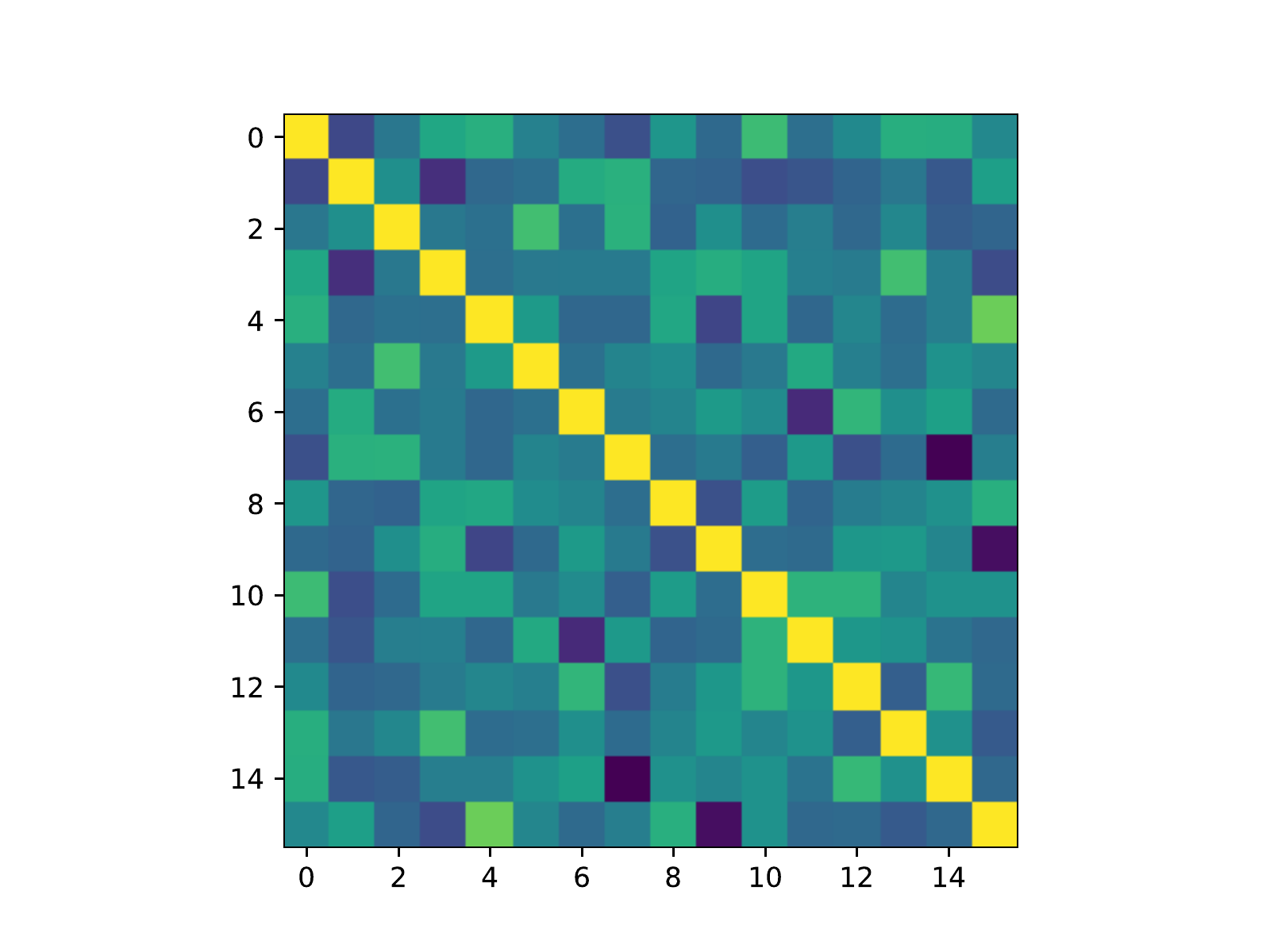}}
\end{tabular}
\end{center}
\vspace{-.25in}
\caption{Example Channel Correlation Matrices Within each Shift Group}
\vspace{-.05in}
\label{fig:correlation}
\end{figure}
}

\subsection{Normalized Channel Contributions}

We analyze the same $CSC$ module from the section above. Consider the $m$-th channel's contribution to the output of the module, we take the norm of the $m$-th row from the point-wise convolution kernel as $\|P_{m, :}\|_2$ for an approximation of ``contribution''. We compute the contribution of all 144 channels in the given $CSC$ module, normalize their contribution and plot it in Figure~\ref{fig:NCC}. Note that the shift contributions are anisotropic, and the largest contributions fall into a cross, where horizontal information is accentuated. This suggests that better heuristics for allocating channels among shift groups may yield neural networks with even higher per-FLOP and per-parameter accuracy.

\section{Conclusion}
We present the shift operation, a zero-FLOP, zero-parameter, easy-to-implement alternative to convolutions for aggregation of spatial information. To start, we construct end-to-end-trainable modules using shift operations and pointwise convolutions. To demonstrate their efficacy and robustness, we replace ResNet's convolutions with shift modules for varying model size constraints, which increases accuracy by up to 8\% with the same number of parameters/FLOPs and recovers accuracy with a third of the parameters/FLOPs. We then construct a family of shift-based neural networks. Among neural networks with around 4 million parameters, we attain competitive performance on a number of tasks, namely classification, face verification, and style transfer. In the future, we plan to apply ShiftNets to tasks demanding large receptive fields that are  prohibitively expensive for convolutions, such as 4K image segmentation.

\section*{Acknowledgement}
This work was partially supported by the DARPA PERFECT program, Award HR0011-12-2-0016, together with ASPIRE Lab sponsor Intel, as well as lab affiliates HP, Huawei, Nvidia, and SK Hynix. This work has also been partially sponsored by individual gifts from BMW, Intel, and the Samsung Global Research Organization. We thank Fisher Yu, Xin Wang for valuable discussions. We thank Kostadin Ilov for providing system assistance. 






\pagebreak
{\small
\bibliographystyle{ieee}
\bibliography{egbib}
}

\end{document}